\documentclass[10pt,twocolumn,letterpaper]{article}

\usepackage{cvpr}              %

\usepackage[dvipsnames]{xcolor}

\definecolor{cvprblue}{rgb}{0.21,0.49,0.74}
\usepackage[pagebackref,breaklinks,colorlinks,citecolor=cvprblue]{hyperref}
\usepackage{graphicx}
\usepackage{amsmath}
\usepackage{mathtools}
\usepackage{pbalance}
\def\paperID{15} %
\def\confName{CVPR}
\def\confYear{2024}

\title{Towards Practical Single-shot Motion Synthesis}

\author{Konstantinos Roditakis
\quad \quad
Spyridon Thermos
\quad \quad
Nikolaos Zioulis
\\
\\
Moverse
\\
\
{\tt\small \{kostas,spiros,nick\}@moverse.ai}
}

\begin{document}
\twocolumn[{%
\renewcommand\twocolumn[1][]{#1}%
\maketitle

\begin{center}
    \centering
    \includegraphics[width=\textwidth]{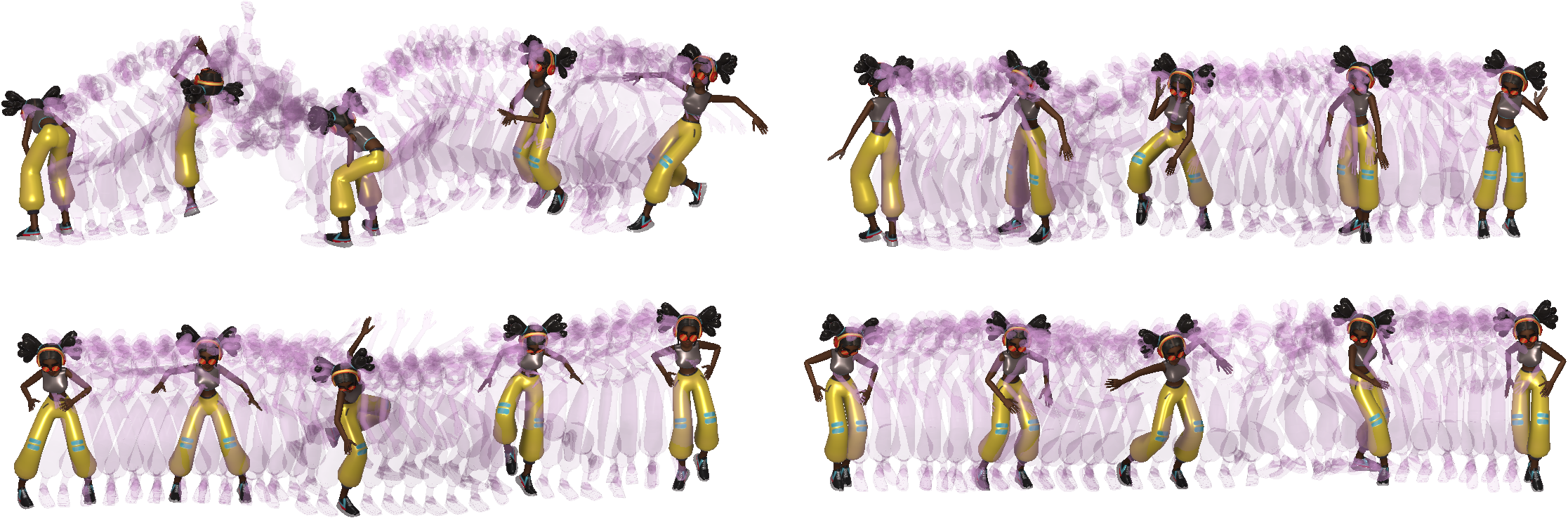}
    \captionof{figure}{
        Given a single motion sequence our improved GAN learns to generate motion variations in minutes using mini-batch training and transfer learning, without compromising the quality or the diversity of synthesized motion. Here we generate variations of the Mixamo sequences (raw-major order): a) ``breakdance freezes" , b) ``dancing", c) ``swing dancing, and d) ``salsa dancing" .
    }
    \label{fig:teaser}
\end{center}
}]
\begin{abstract}
Despite the recent advances in the so-called ``cold start" generation from text prompts, their needs in data and computing resources, as well as the ambiguities around intellectual property and privacy concerns pose certain counterarguments for their utility. 
An interesting and relatively unexplored alternative has been the introduction of unconditional synthesis from a single sample, which has led to interesting generative applications. 
In this paper we focus on single-shot motion generation and more specifically on accelerating the training time of a Generative Adversarial Network (GAN).
In particular, we tackle the challenge of GAN's equilibrium collapse when using mini-batch training by carefully annealing the weights of the loss functions that prevent mode collapse. 
Additionally, we perform statistical analysis in the generator and discriminator models to identify correlations between training stages and enable transfer learning.
Our improved GAN achieves competitive quality and diversity on the Mixamo benchmark when compared to the original GAN architecture and a single-shot diffusion model, while being up to $\times 6.8$ faster in training time from the former and $\times 1.75$ from the latter.
Finally, we demonstrate the ability of our improved GAN to mix and compose motion with a single forward pass. 
Project page available at \url{https://moverseai.github.io/single-shot}
\end{abstract}
    
\section{Introduction}
\label{sec:intro}

Since the advent of the Large Language Models (LLMs), the so-called ``cold-start" generation has attracted impressive attention as a viable path to artificial general intelligence. 
In fact, LLMs have demonstrated proficiency in various domains~\cite{damonlpsg2023videollama, blip2} transforming text information to images, scenes, and more recently to human pose~\cite{feng2023posegpt} and motion~\cite{jiang2023motiongpt} even coupled with denoising diffusion model~\cite{ho2020ddpm} variants~\cite{kim2022flame,tevet2023human}.
However, the said models require massive computational resources and are trained on vast amounts of annotated data, which can include personal and sensitive information.
Their lack of interpretability and explainability~\cite{llm2024explain} poses a certain risk that they may inadvertently memorize and reproduce this sensitive data during generation, which raises ethical and intellectual property barriers.
Inference, or using a pre-trained LLM for generating text, also demands significant computational resources.
LLMs may reflect and potentially amplify biases present in the training data, leading to biased or unfair outputs. 
Addressing bias requires careful curation of training data and model design, which can be resource-intensive and challenging.

An interesting alternative to cold-start generation and its challenges are the single sample generative models, which can serve as powerful editing tools as they provide a good balance between pluralism and context preservation. 
Pioneered in the domain of images, they have been used to remap \cite{shocher2019ingan}, composite and edit \cite{rottshaham2019singan} images, increase their resolution \cite{shocher2018zero}, and also generate \cite{jetchev2016nipsw} and expand \cite{zhou2018non} textures.
Even though they are very important as they can help overcome intellectual property and data privacy/sensitivity issues, they still remain a relatively unexplored topic.
In the context of 3D content, they are even more important as -- compared to text, images, and/or video -- 3D data are more challenging to acquire in quantity.
Specifically for 3D motion, two variants have been recently introduced, GANimator \cite{li2022ganimator} and SinMDM \cite{raab2024single} that represent the two dominant classes of approaching this task, namely using generative adversarial networks (GANs) \cite{goodfellow2020generative} and denoising diffusion probabilistic  models (DDPMs) \cite{ho2020denoising}.
While both are hyper-parameter and architecture sensitive approaches, the latter (SinMDM) was shown to be faster to train than the former (GANimator), as well as support more applications without re-training.
On the other hand, GANimator relies on a single forward pass, and thus, exhibits much faster inference performance compared to the iterative nature of diffusion.
Content editing applications need to support interactive workflows (\textit{i.e.}~real-time inference) but at the same time the workflows are on-demand, which also imposes constraints on the set up time (\textit{i.e.}~fast training).

In this work we focus on improving the training time of GANimator that already delivers real-time inference, taking a step towards the practical realization of single motion generative editing.
We find that a major limitation of single sample GANs compared to DDPMs is the lack of mini-batch training.
The latter is a challenge for single sample training, especially in the unconditional case, a task that needs to balance adversarial training with a latent anchoring objective. 
Further, we show that the hierarchical nature of GANimator is an unexplored trait that can be exploited to improve training time and realize more editing applications in a unified manner and without retraining.

Summarizing, our contribution is two-fold:
\begin{itemize}
    \item We study the challenges for mini-batch and hierarchical training in the single sample GAN regime and show increase its performance by a factor of 10 through combining mini-batch training and cross-stage transfer learning.
    Our GANimator variant trains faster than SinMDM and simultaneously offers real-time inference performance.
    \item We show how we can realize diverse motion compositing in a single forward pass by exploiting the hierarchical nature of GANimator, expanding its application domain.
\end{itemize}

\section{Prior Art}
\label{sec:related}

\noindent\textbf{Data-driven motion generation.}
Synthesizing human motion is a long standing problem with some of the early approaches being based on statistical modelling \cite{bowden2000learning}, exemplar \cite{arikan2002interactive,arikan2003motion} or graph walk \cite{kovar2023motion} based composition, and learning the distribution of novel constructs like motion textons \cite{li2002motion}.
Given larger datasets and modern learning techniques it is now possible to generate motion from text \cite{petrovich2022temos,tevet2023human,azadi2023make}, offer a GPT-like interface for motion \cite{jiang2023motiongpt} and use said models to perform editing tasks \cite{shafir2023human,athanasiou2022teach}.
More general motion synthesis and editing frameworks have also been presented \cite{holden2016deep} due to the wider availability of large scale motion datasets.
Still, large dataset acquisition is challenging, especially when considering text prompt annotations.
It also comes with barriers related to the sensitivity of the data either from a personal or creative point of view.

\noindent\textbf{Single-shot generation.}~Simple sample generative models are a promising alternative to overcome these barriers as they train specialized models that generate variations of a specific sample only.
InGAN \cite{shocher2019ingan} first showed that it is possible to train a conditional GAN model on a single image for the task of remapping, with follow-up works focusing on texture generation and editing \cite{jetchev2016nipsw,bergmann2017icml,li2016patch}.
Following a progressive learning scheme across multiple stages, each operating on a different scale, SinGAN \cite{rottshaham2019singan} is the first unconditional GAN trained on a single image.
Crucially it relies on a patch-based discriminator \cite{li2016patch,isola2017cvpr} restricting its receptive field, which is coupled with a reconstruction objective that anchors its latent space protect the model training from a mode collapse.

Nonetheless, this increases the training time, which -- when considering the single sample context -- is a big obstacle for practical application use.
Different variants followed, with ExSinGAN \cite{zhang2021exsingan} using external priors to improve structural and semantic performance of the model, while ConSinGAN \cite{hinz2021improved} focused on improving the training time by training stages in parallel and reducing their number.
To improve the preservation of sample's context, OneShotGAN \cite{sushko2021one} introduced a dual discriminator to supervise both the global context, as well as the patch-based layout.
PetsGAN \cite{zhang2022petsgan} improves training time by leveraging external priors whereas HP-VAE-GAN \cite{gur2020hierarchical} opts for a hybrid VAE-GAN scheme to enable single video generation.
The challenge of quickly training single sample generative models mostly stems from the nature of the adversarial game, and thus, novel approaches that reformulate the task to a reconstruction \cite{yoo2021sinir} or nearest neighbor retrieval \cite{granot2022drop} one manage to greatly accelerate training time, but at the expense of generation variance.
Using diffusion models, like SinDDM \cite{kulikov2023sinddm}, is another alternative to reducing training time due to mini-batch training.
The interaction between the added reconstruction objective with the adversarial game is difficult to balance, and is the reason why single sample GANs are typically trained with a batch size of one as training destabilizes when increasing the batch size.

\noindent\textbf{Single-shot 3D content generation.}
More recently, scarce efforts have been made to demonstrate single sample generation to 3D content.
Sin3DM \cite{wu2023sin3dm} trains a diffusion model on a single sample, and leverages and intermediate latent representation to overcome memory and runtime performance issues.
SinGRAF \cite{son2023singraf} bridges a neural rendering representation to accomplish single sample variation generation of specific 3D scenes.
For 3D motion single sample generation there exist two approaches, GANimator \cite{li2022ganimator} and SinMDM \cite{raab2024single}, using adversarial training and denoising diffusion respectively.
GANimator consists of 7 stages of skeleton-aware convolutions~\cite{aberman2020skeleton} forming 4 pyramid levels (2-2-2-1 stages) emulating the pyramidal design of SinGAN in the temporal domain, where each stage learns to generate a sequence of different length (up to the input's one).
Changing the core of the aforementioned works, Raab~\etal~\cite{raab2024single} present a motion diffusion model that learns to generate variations of a single motion sequence. 
SinMDM follows the structure of the UNet-based diffusion model presented in~\cite{nichol2022icml} but with a significant detail; they add shift-invariant local attention layers~\cite{arar2022cvpr} to decrease the receptive field of UNet and avoid overfitting in the single sample scenario.
GANimator is slower to train but significantly faster when generating samples, whereas SinMDM exploits mini-batch training to reduce training time but requires an iterative diffusion process at inference.
Further, SinMDM supports more applications without requiring retraining, an important advantage from a practical point of view.

\section{Efficient Novel Motion Synthesis}
\label{sec:method}

In this section we present our findings for improving the performance of single-shot motion synthesis in more detail, starting with enabling mini-batch training, followed by transfer learning between stages.

\subsection{Background}
\label{sec:background}
We present the background of GANimator~\cite{li2022ganimator} and formalize the notations to set the stage for our improvements in the GAN training process. 

\textbf{Data representation.}
Li~\etal~\cite{li2022ganimator} form a motion representation $\mathcal{M}_{T} \equiv \mathbb{R}^{T\times(JQ+C+3)}$, where $T$ indicates the number of frames, $J$ is the number of skeleton joints, $Q=6$ corresponds to the 6D rotation representation of the joints, and $C$ indicates the foot contact labels followed by the 3D representation of the root joint including the $x$- and $z$-axis velocity and the $y$-axis position.

\begin{figure}
    \centering
    \includegraphics[width=\columnwidth]{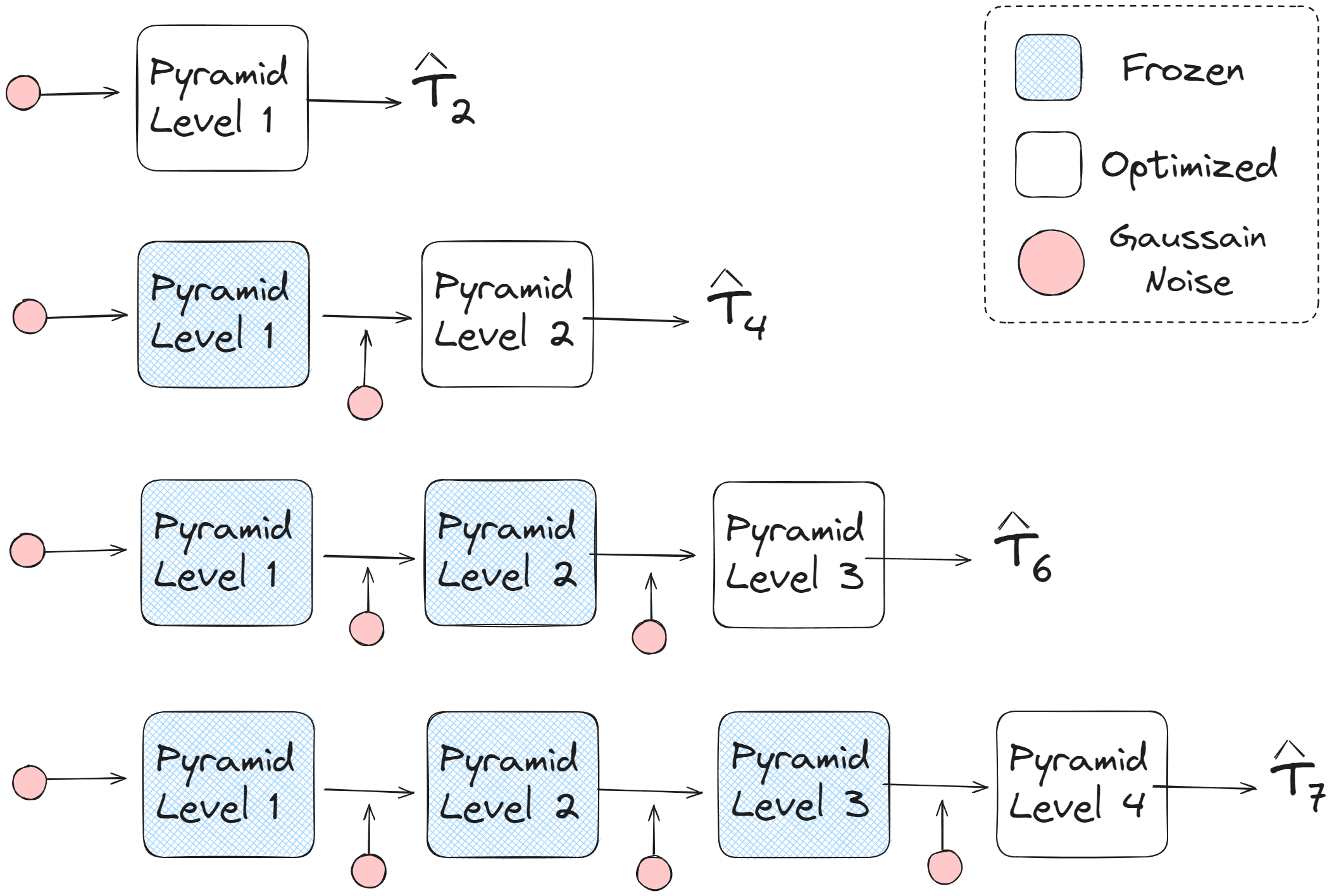}
    \caption{A schematic representation of the GANimator~\cite{li2022ganimator} gradual training architecture. From top to bottom, each pyramid level is learning to generate motion features at time scale ($\hat{T}_{*}$). When a pyramid level is trained it serves as a frozen feature extractor for the next level. Note that the last pyramid level ($L_{4}$) consists of only one $\{G(\cdot),D(\cdot)\}$ pair.}
    \label{fig:full_stages_ganimator}
    \vspace{-0.1in}
\end{figure}

\begin{figure}
    \centering
    \includegraphics[width=\columnwidth]{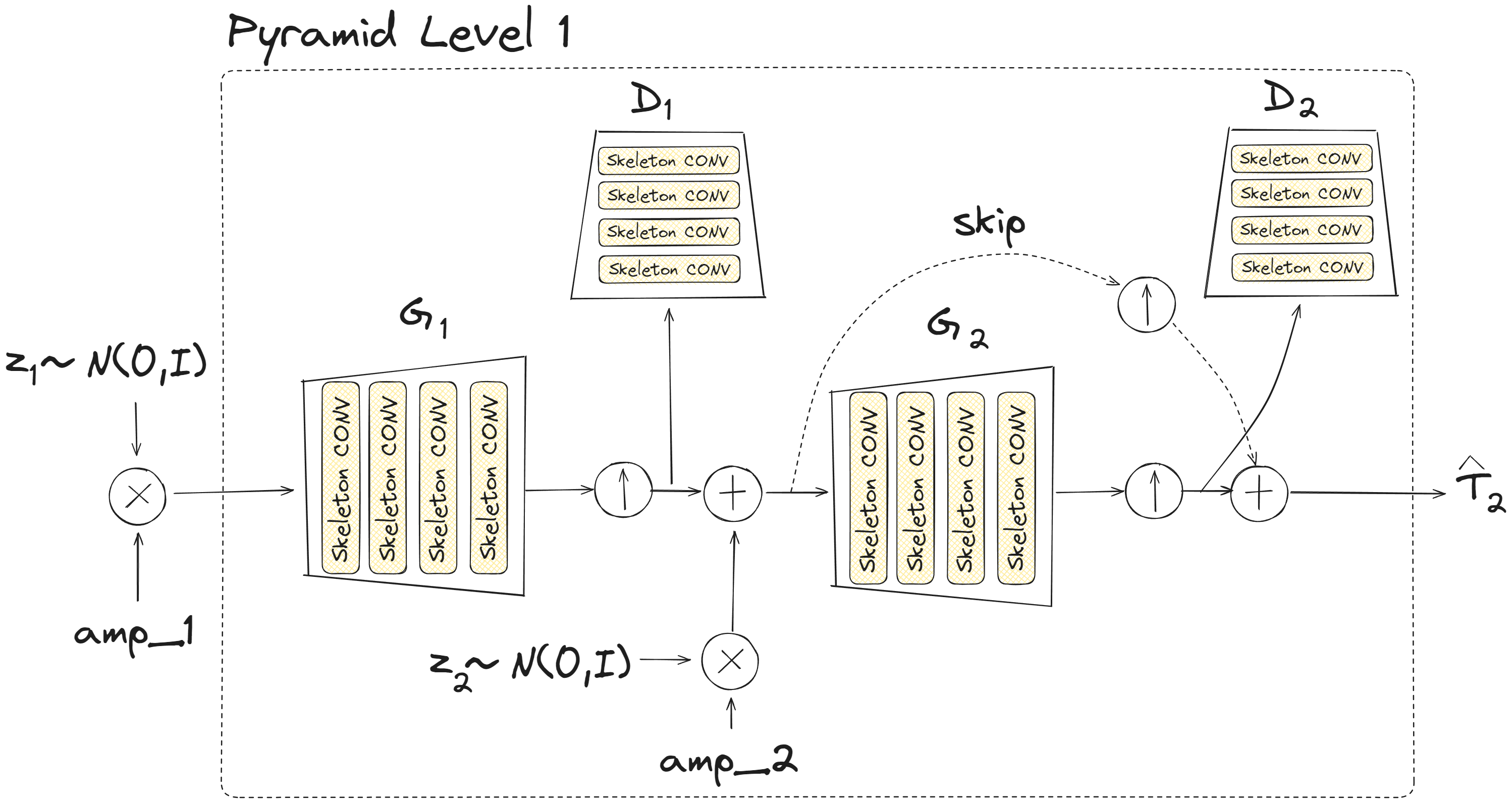}
    \caption{GANimator's pyramid level 1 in detail: $G_{1}$ is responsible for learning the mapping between sampled noise $z_{1}\sim\mathcal{N}(0,I)$ (multiplied by a predefined amplitude) and the motion representation $T$, which is then upsampled ($\hat{T}_{1}$) and sent to $D_{1}$; before $G_{2}$ we add extra noise to the predicted motion features $\hat{T}_{}1$ to force the model to learn variations of the input sample.}
    \label{fig:train_stage_ganimator}
\end{figure}

\textbf{GAN architecture.}
GANimator follows a coarse-to-fine motion feature learning approach, with $S$ stages of generators $G(\cdot)$ and discriminators $D(\cdot)$ pairs. 
The GAN model does not train all stages in an end-to-end manner, but follows a gradual learning approach, as shown in Fig.~\ref{fig:full_stages_ganimator}.
The stages are groups of pairs of $\{G(\cdot),D(\cdot)\}$ forming $4$ pyramid levels $L_{*}$ - except for $L_{4}$ that contains only $S_{7}$.
Each level's stage learns to generate the motion features of increasing temporal resolution. 
For the rest of the paper we use the subscript to denote stage levels and the superscript to denote the pyramid levels (\eg $\hat{T}^{1}_{2}$ corresponds to the generated motion features from the second $G(\cdot)$ of the first pyramid level). 
Each generator $G(\cdot)$ and discriminator $D(\cdot)$ consists of 4 skeleton-aware convolutions~\cite{aberman2020skeleton} followed by leaky ReLU activations (except for the final convolution layer). 
As depicted in Fig.~\ref{fig:train_stage_ganimator}, $G^{1}_{1}$ is responsible for learning the mapping between the sampled noise and the motion representation denoted as \textcolor{red}{$\hat{T}+{1}$}, \ie $\hat{T}^{1}_{1} = G^{1}_{1}(z_{1})$, while the rest $G^{*}_{\{2,\dots, S\}}(\cdot)$ form a hierarchical auto-regressive process that progressively upsamples the generated sequence:
\begin{equation}
\label{eq:total_loss}
    \hat{T}^{\ell}_{i} = G^{\ell}_{i}(\hat{T}^{\ell}_{i-1}, z_{i}), \ i \in \{2,\dots, S\}, \ \ell \in \{1,\dots, L\},
\end{equation}
where $z_{i}$ is sampled from an i.i.d.~normal distribution $\mathcal{N}(0, I)$ and multiplied with an decreasing amplitude $\sigma_{i}$. 

\textbf{Losses.}
The model is trained with multiple losses for securing an equilibrium in the adversarial game between generators and discriminators, having the constraint that there is only a single sample to train the model. 
Both $G(\cdot)$ and $D(\cdot)$ are supervised with the Wasserstein variant from~\cite{gulrajani2017neurips}:

\begin{equation*}
    \label{eq:wasser_loss}
    \begin{multlined}
        \mathcal{L}_{adv} = \mathbb{E}_{\hat{T}_{i}\sim\mathbb{P}_{G_{i}}} [D_{i}(\hat{T}_{i})] - D_{i}(T_{i})\\ 
        \notag+ \ \lambda_{gp}\mathbb{E}_{\tilde{T}_{i}\sim\mathbb{P}_{G_{i}}} \Bigl[ \Bigl( || \nabla D_{i}(\tilde{T}_{i}) ||_{2} - 1 \Bigr)^{2} \Bigr],
    \end{multlined}
\end{equation*}
where $\mathbb{P}$ denotes a learned distribution and $\tilde{T}_{i} = \alpha\hat{T}_{i} + (1-\alpha)T_{i}$ is a linear combination of the generated and ground truth motion features, respectively. 
The last term of the equation, \ie the gradient penalty regularization, enforces Lipschitz continuity and stabilizes the training. 
To prevent mode collapse due to the single-sample training the $G(\cdot)$ are additionally supervised by an $L1$ reconstruction loss:
\begin{equation}
    \mathcal{L}_{rec} = || G_{i}(\hat{T}_{i-1}, z^{*}_{i})-T_{i} ||_{1},
\end{equation}
where $z^{*}_{i}$ denotes a predefined noise code (at different levels of amplitude $i \in \{1,\dots,S\}$) that is set to approximate the reconstruction of $T$. 
Lastly, a regularization loss is added to supervise the contact of the predefined set of joints, denoted as $\mathcal{C}$, with the ground:
\begin{equation}
    \mathcal{L}_{con} = \frac{1}{T|\mathcal{C}|}\sum_{j\in\{\mathcal{C}\}}\sum^{T}_{t=1} || \mathcal{V}^{t,j} ||^{2}_{2} \cdot \mathrm{S}(\mathcal{C}^{t,j}),
\end{equation}
where S denotes the skewed Sigmoid function that forces the output to be almost binary, and $\mathcal{V}$ is the velocity computed for the joints of $\mathcal{C}$ using forward kinematics.

Although each loss operates on a different part of the GAN training process, we can summarize them as:
\begin{equation}
    \mathcal{L} = \lambda_{adv}\mathcal{L}_{adv} + \lambda_{rec}\mathcal{L}_{rec} + \lambda_{con}\mathcal{L}_{con},
\end{equation}
where $\lambda_{adv}, \lambda_{rec}, \lambda_{con}$, and $\lambda_{GP}$ are responsible for weighting the contribution of each loss to the adversarial game of the single sample learning. Li~\etal~\cite{li2022ganimator} discuss the contribution of each loss to the final result, while our focus is concentrated in the balancing between the losses.

\subsection{Mini-batch Training} 
One major advantage of the single-shot diffusion models vs single-shot GANs is the exploitation of mini-batch training. 
In fact, finding the equilibrium in the adversarial game of a single-shot GAN is challenging and the mode collapse is a common result when trying to set the batch size larger than $1$.
As discussed in~\cite{salimans2016neurips}, in a data-driven adversarial game mini-batching helps the discriminator to understand when the generator produces samples of very low variation and avoid mode collapse due to this side information. 
In single-shot generation, the outputs of the generator are by definition minor variations of the single input sample, while the narrow receptive field of the patch-based discriminator and the reconstruction loss $\mathcal{L}_{rec}$ are responsible for preventing the mode collapse. 
Naturally, increasing the batch size counters the intuition that the narrow receptive field of the discriminator will prevent mode collapse, while the reconstruction loss tends to reduce the coverage by forcing the average of the batch to be similar to the input sequence.
Since mini-batching critically improves the training time, we performed an ablation study on the weights of $\mathcal{L}_{rec}$ and $\mathcal{L}_{adv}$ in a quest for finding the correct combination that preserves the equilibrium in the adversarial game. However, note that this is not a straight-forward weight tuning process since each loss operates on different parts of the training process (\ie different optimizer).
Starting from the original weight values $\lambda_{adv}=1$ and $\lambda_{rec}=50$ we choose a stage-based linear annealing of: a) $\lambda_{adv}$, b) $\lambda_{rec}$, c) both $\lambda_{adv}$ and $\lambda_{rec}$. 
As shown in Table~\ref{tab:ablation}, we achieve the best results with (c) when we boost $\mathcal{L}_{adv}$ in the early stages where mapping from sampled noise to motion representation takes place, while the later stages are dominated by the reconstruction loss to ensure that no rare mini-motions are omitted.

\begin{figure}
    \centering
    \includegraphics[width=0.95\columnwidth]{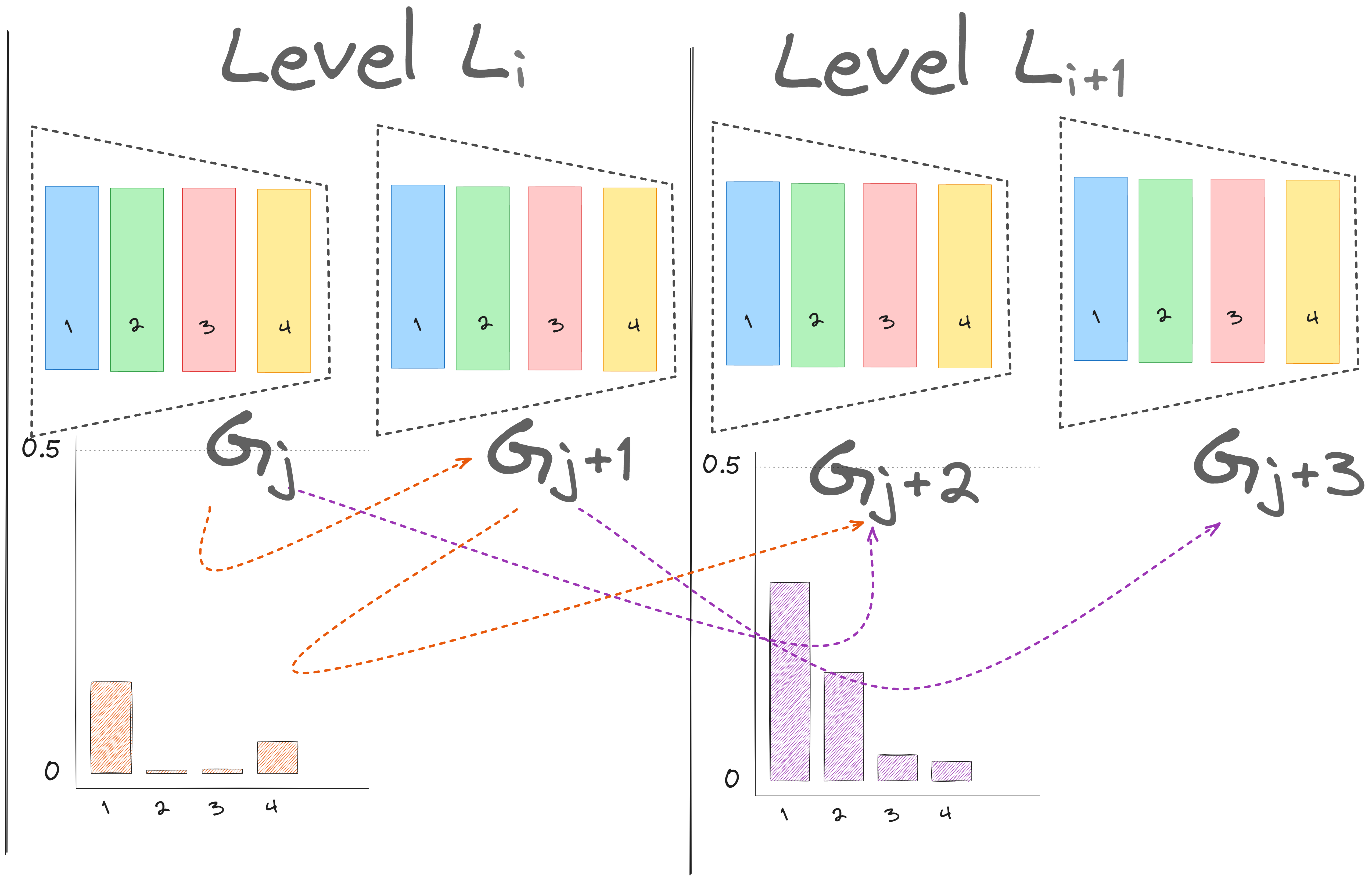}
    \caption{Representation similarities across stages and levels. 
    Each generator ($G_{j}$) exhibits low representation similarity  scores with the following generator ($G_{j+1})$ (\textcolor{orange}{orange dashed arrows \& plots}).
    Yet, we find that the corresponding stages' generators ($G_j \longleftrightarrow G_{j+2}$) across levels ($L_i$) exhibit higher similarity scores for the early layers only ($1 \, \& \, 2$ -- \textcolor{violet}{purple dashed arrows \& plots}). 
    Transferring the trained generators' ($G_j$) layers' weights to the  next level's ($L_{i+1}$) corresponding stages' generators ($G_{j+2}$) before training them, improves convergence rate. }
    \label{fig:correlations}
\end{figure}

\begin{table*}[]
\caption{Ablation study of the model parameters for the GANimator~\cite{li2022ganimator} architecture. The demonstrated results correspond to motion ``salsa dance" of the Mixamo benchmark. Although the GAN variant with batch size 24 (Abl \#10) is the fastest to train, Abl \#9 exhibits the best trade-off between quality/diversity and train time.}
\label{tab:ablation}
\vspace{-0.1in}
\resizebox{\textwidth}{!}{%
\begin{tabular}{lcccccccccccc}
 & \multicolumn{6}{c}{Quality \& Diversity} & \multicolumn{5}{c}{Model Parameters} & Performance \\ \cline{2-13} 
 & Coverage $(\uparrow)$ & Global Div. $(\uparrow)$ & Local Div. $(\uparrow)$ & SIFID $(\downarrow)$ & Inter Div. $(\uparrow)$ & Intra Div. $(\downarrow)$ & BS & Iters & l\_adv & l\_rec & \multicolumn{1}{l|}{TL} & Train Time $(\downarrow)$ \\ \hline
Baseline & 1.00 & 1.12 & 1.05 & 3.47 & 1.67 & 1.72 &  1 & 105k & 1 & 10 & \multicolumn{1}{l|}{} & 5h30m \\
Abl \#1 & 1.00 & 0.98 & 0.92 & 3.61 & 1.19 & 1.90 & 1 & 210k & 1 & 10 & \multicolumn{1}{l|}{} & 10h21m \\
Abl \#2 & 1.00 & 0.86 & 0.81 & 4.49 & 0.73 & 2.37 & 1 & 105k & 5 & 10 & \multicolumn{1}{l|}{} & 5h30m \\
Abl \#3 & 1.00 & 0.71 & 0.67 & 3.41 & 0.47 & 1.87 & 1 & 210k & 5 & 10 & \multicolumn{1}{l|}{} & 10h21m \\
Abl \#4 & 0.96 & 1.51 & 1.41 & 4.66 & 1.80 & 2.48 & 16 & 105k & 5 & 10 & \multicolumn{1}{l|}{} & 0h52m \\
Abl \#5 & 1.00 & 1.31 & 1.22 & 3.36 & 1.72 & 2.31 & 16 & 210k & 5 & 10 & \multicolumn{1}{l|}{} & 1h47m \\
Abl \#6 & 1.00 & 1.26 & 1.18 & 3.99 & 1.99 & 2.26 & 16 & 210k & 5 & 100 & \multicolumn{1}{l|}{} & 1h47m \\
Abl \#7 & 1.00 & 1.34 & 1.27 & 4.59 & 2.17 & 2.70 & 16 & {[}210, 210, 105, 52.5{]} & 5 & 100 & \multicolumn{1}{l|}{y} & 0h42m \\
Abl \#8 & 1.00 & 1.43 & 1.34 & 4.20 & 1.95 & 2.23 & 16 & 210k & {[}5.0,5.0,2.5,1.0{]} & {[}50,75,100,100{]} & \multicolumn{1}{l|}{} & 1h47m \\
Abl \#9 & \textbf{1.00} & \textbf{1.41} & \textbf{1.32} & \textbf{3.93} & \textbf{1.93} & \textbf{2.30} & \textbf{16} & {[}\textbf{210, 210, 105, 70}{]} & \textbf{{[}5.0,5.0,2.5,1.0{]}} & \textbf{{[}50,75,100,100{]}} & \multicolumn{1}{l|}{\textbf{y}} & \textbf{0h48m} \\
Abl \#10 & 0.96 & 1.58 & 1.48 & 4.98 & 2.07 & 2.32 & 24 & 105k & 5 & 10 & \multicolumn{1}{l|}{} & 0h29m \\
Abl \#11 & 0.99 & 1.43 & 1.33 & 5.43 & 1.76 & 2.38 & 24 & 210k & 5 & 100 & \multicolumn{1}{l|}{} & 0h57m \\ \hline
\end{tabular}%
}
\end{table*}

\subsection{Cross-Stage Transfer Learning}
Apart from increasing convergence through larger mini-batch training, the hierarchical structure of single sample GANs can be exploited to improve convergence. 
ConSinGAN \cite{hinz2021improved} trained multiple stages in parallel and re-used the discriminator from the previous stage to improve performance by transferring its weights to initialize the next stage discriminator.
GANimator \cite{li2022ganimator} already trains two stages in the same pyramid level but does not perform discriminator transfer across stages or pyramid levels, even though the same discriminator is used -- in terms of architecture and capacity - as in ConSinGAN and in contrast to SinGAN that increases discriminator capacity at each stage.

Curiously, re-using the generator weights does not lead to improved performance.
To study this we turn to studying the neural network representations using finer grained layer-wise analysis. 
It has been shown that the similarity of representations across layers can be measured despite the higher dimensionality of the representations \cite{kornblith2019similarity}.
We perform linear centered kernel alignment (CKA) for all stage combinations across all pyramid levels.

Our results, depicted in Fig.~\ref{fig:correlations}, show that despite the hierarchical approach of using the same model and capacity for each stage, most layers exhibit low similarity scores.
Yet we find that the early layer representations across the stages of each pyramid level, exhibit higher levels of similarity.
Contrary to ConSinGAN that outputs features at each level apart from the last, GANimator reconstructs the output motion at each level.
Therefore the early convolutional layers operate on similar features and, as indicated by their CKA scores, extract similar representations, whereas the latter layers apply the level's motion motifs, style and details. Based on this analysis, we design a generator transfer learning scheme across levels and stages where each generator's ($G^i_j$) early layer weights are initialized from the early layer weights of the previous level generator ($G^{i-1}_{j-2}$).
This ensures that each next training stage is closer to the converged state allowing us to reduce the number of iterations significantly, further boosting training time.

\begin{table*}[]
\footnotesize
\caption{Quality, diversity and performance comparison between our improved GAN, GANimator~\cite{li2022ganimator} and SinMDM~\cite{raab2024single}. 
We present the average results on the Mixamo benchmark. 
The presented train and inference times are measured on a NVIDIA RTX 3060 GPU on that ``salsa dance" with character ``Joe" sample from Mixamo. 
Following~\cite{raab2024single} we compute the harmonic mean of the 6 metrics to provide a balanced overall result for each experiment.}
\label{tab:comparison}
\resizebox{\textwidth}{!}{%
\begin{tabular}{lccccccccc}
 & \multicolumn{7}{c}{Quality \& Diversity} & \multicolumn{2}{c}{Performance} \\ \cline{2-10} 
 & \multicolumn{1}{l}{Coverage $(\uparrow)$} & \multicolumn{1}{l}{Global Div. $(\uparrow)$} & \multicolumn{1}{l}{Local Div. $(\uparrow)$} & \multicolumn{1}{l}{SIFID $(\downarrow)$} & \multicolumn{1}{l}{Inter Div. $(\uparrow)$} & \multicolumn{1}{l}{Intra Div. $(\downarrow)$} & \multicolumn{1}{l|}{Harm. Mean $(\uparrow)$} & \multicolumn{1}{l}{Train Time $(\downarrow)$} & \multicolumn{1}{l}{Inf. Time $(\downarrow)$} \\ \hline
\cite{li2022ganimator} & 0.95 & 1.02 & 0.96 & 1.15 & 1.49 & 2.12 & \multicolumn{1}{c|}{0.50} & 5h30m & \textbf{5.2ms} \\
\cite{raab2024single} & 0.94 & 1.42 & 1.00 & 1.08 & 1.43 & 1.93 & \multicolumn{1}{c|}{0.58} & 1h24m & 5000ms \\
Ours & 0.89 & 1.46 & 1.35 & 1.99 & 1.75 & 1.57 & \multicolumn{1}{c|}{0.52} & \textbf{0h48m} & \textbf{5.2ms} \\ \hline
\end{tabular}%
}
\end{table*}

\section{Experiments}
\label{sec:experiments}
Our experimental section is split into the quantitative analysis, discussing the metrics of the literature and comparing our improved approach with the state-of-the-art (SoTA), and the qualitative analysis that consists of new applications for single-shot GAN-based generation.

\begin{figure*}
    \centering
    \includegraphics[width=\textwidth]{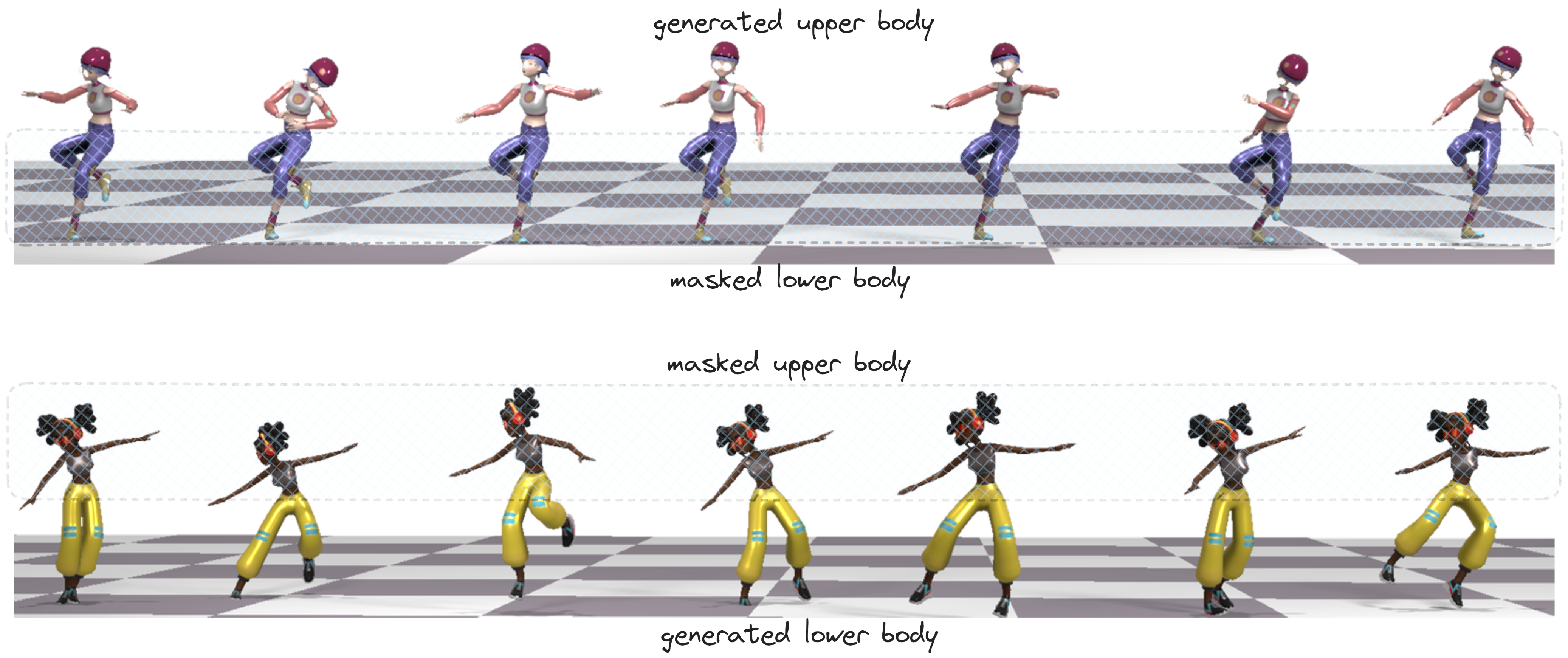}
    \caption{Body-part composition: (top) snapshots of 7 generated ``swing dancing" motion variants with the lower body masked (shaded area) to preserve the original sequence pose, while the upper body is randomly generated; (bottom) snapshots of 7 generated ``swing dancing" motion variants with the upper body masked to preserve the original sequence pose, while the lower body is randomly generated. Note that the root joint is considered as part of the lower body, thus the top samples retain the global orientation of the original motion.}
    \label{fig:body_part_composition}
\end{figure*}

\subsection{Quantitative analysis}
Fist, we briefly report the metrics used in the literature for assessing the quality and the diversity of single-sample generation, present our implementation details, and discuss our results compared to similar models~\cite{li2022ganimator,raab2024single}. 

\textbf{Metrics.}~For a fair comparison we use the metrics presented in~\cite{li2022ganimator,raab2024single}, which try to measure the local and global diversity of the generated sequences, as well as their quality in terms of plausibility and coverage. 
We give a brief description for each metric, commenting on its usefulness.

\textbf{Quality:} The combination of the coverage and the plausibility expressed as distance from a distribution form a robust pair for understanding how realistic is the generated motion. 
Li~\etal~\cite{li2022ganimator} consider a temporal window $T_{w}$ of the input sequence covered if its distance from its nearest neighbor $Q_{w}$ in the generated sequence is less than a predefined threshold $\epsilon$; on the other hand, SinMDM adopts the Fr\'echet Inception Distance (FID) variant from~\cite{rottshaham2019singan}, which uses the deep features of an earlier convolutional layer of the Inception Network~\cite{szegedy2015inception} in order to compute the FID statistics between the input and the generated sequences.
As noted in~\cite{raab2024single}, coverage has been experimentally shown to be sensitive, thus we choose to interpret it jointly with the single sample FID (SiFID) to truly describe quality.

\textbf{Diversity:} \cite{li2022ganimator} and \cite{raab2024single} compute the local and global diversity of the generated sequence following different approaches; Li~\etal~\cite{li2022ganimator} use high-level features (\ie rotation angles) to compute distances either from the nearest neighbors of the input sequence, while Raab~\etal~\cite{raab2024single} use embeddings of motion features from a pre-trained motion encoder for computing the corresponding distances.
From our experiments, we confirm the superiority of deep features over raw input features~\cite{zhang2018cvpr} in interpreting the inter-diversity and the intra-diversity of the generated sequences.
However, we run our evaluation using all presented metrics, as well as the harmonic mean from~\cite{raab2024single} that attempts to describe both quality and diversity with one value.

\textbf{Data.}~We use the Mixamo\footnote{\url{https://www.mixamo.com}} sequences presented in \cite{raab2024single} to evaluate our improved GAN against the GANimator and the SinMDM in terms of training time, inference time, quality and diversity using the aforementioned metrics.

\textbf{Implementation details.}~We use the provided PyTorch~\cite{paszke2019pytorch} implementations for GANimator and SinMDM and adopt the pretrained motion encoder from the SinMDM codebase to extract the motion embeddings for SiFID and inter/intra diversity metrics computation. 
All experiments are conducted on a NVIDIA RTX 3060 GPU.

\begin{figure*}
    \centering
    \includegraphics[width=\textwidth]{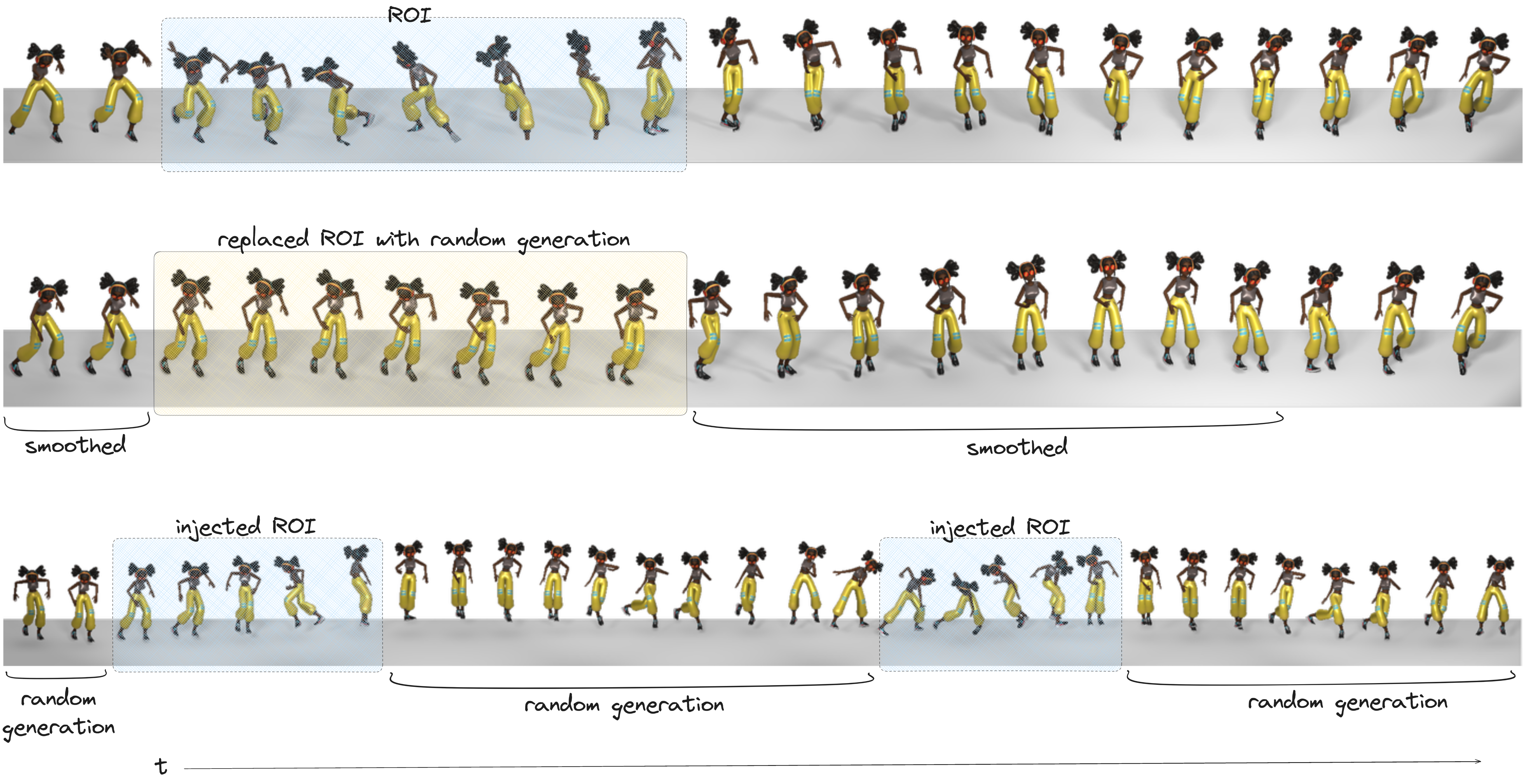}
    \caption{Full-body motion composition example: (top) we select the region of interest (ROI), \ie a clock-wise spin, from the ``salsa dancing" sequence of the Mixamo corpus; (middle) a variat of the original sequence is generated with the selected ROI removed and the temporal space being ``filled" with motion features generated from noise; (bottom) a new ``salsa dancing" variant is composed with 2 spins being placed at user-selected time steps - the non-ROI region are generated by sampled noise. We depict more frames of the composed ``salsa" variant (bottom) to demonstrate the 2 spins in the same row.}
    \label{fig:composition}
\end{figure*}

\textbf{Results.} 
Table~\ref{tab:ablation} details the performed ablation study for improving the training performance of GANimator by validating the two presented techniques, \ie mini-batch training with equilibrium preservation and transfer learning between pyramid levels. 
As shown in the results, increasing the batch size significantly improves the training time, but should be accompanied with an increase of training iterations, $\lambda_{adv}$, and $\lambda_{rec}$. However, larger batch sizes seem to hurt the $G(\cdot)-D(\cdot)$ adversarial game and leads to inferior performance despite decreasing the training time. 
After having improved training time and demonstrated similar performance with the baseline, we move to exploit the correlations between pyramid levels. 
From Table~\ref{tab:ablation}, we can see that the combination of cross-stage transfer for the generators $G(\cdot)$, and annealing $\lambda_{adv}$ and $\lambda_{rec}$ leads to the best results, while exhibiting the best improvement in training time.
As a next step, we compare our best model with GANimator \cite{li2022ganimator} and SinMDM \cite{raab2024single}. 
From the results in Table~\ref{tab:comparison}, we conclude that our improved GAN achieves slightly better results that its baseline \cite{li2022ganimator} when tested on the Mixamo dataset, while also approaching SinMDM~\cite{raab2024single}. 
However, the results showcase that our GAN exhibits a significant improvement in training time compared to both SoTA (almost $\sim\!\times 7$ and $\sim\!\times 2$ respectively), while being extremely faster in inference time compared to the latter ($\sim\!\times 1000$).

\subsection{Applications}
Apart from the training time performance increase, we introduce new applications that can be performed with a single-shot GAN-based model, such as the body-part and full-body motion composition, in addition to showcasing results on applications introduced by GANimator, like motion re-styling and crowd generation.
Note that contrary to~\cite{li2022ganimator} we focus solely on applications that do not need re-training, as they pose the main challenge and exploit the superiority of GANs over other approaches in terms of performance.

\textbf{Body-part motion composition.}
As detailed in Sec.~\ref{sec:background}, the motion feature $\mathcal{M}_{T}$ of each input sequence $T$ includes a 6D representation about each skeleton joint. 
This allow us to define body binary masks $M$ for the upper and the lower body, which can be applied on the motion features during inference and force the masked area to retain its original values, while the rest of the body's movement will be generated based on randomly sampled noise. 
As depicted in Fig.~\ref{fig:body_part_composition} (top) we use the Mixamo motion ``swing dancing" as input sequence $T$ and we choose to keep the lower-body unaltered, while generating alternative - but natural - versions of the upper-body. 
To achieve that, we use the $G(\cdot)$ trained with this sequence; the hierarchical structure of the model allows us to choose at which level $L_{*}$ to apply the predefined mask $M^{lb}$ that will keep the lower-body (lb) unaltered. 
We choose to apply $M^{lb}$ at $L_{2}$ as it leads to the smoothest blending of the body parts.
Fig.~\ref{fig:body_part_composition} (bottom) demonstrates the application of $M^{ub}$ to the upper-body (ub) of the same motion sample, which preserves its pose despite the generated global rotation and lower body pose. 

\textbf{Full-body motion composition.} 
Assuming a reference motion $T$ of arbitrary length, we consider the following options: a) remove mini-clips of $T$ and use the GAN to ``inpaint" them with generated content, and b) select one (or more) mini-clip(s) from $T$ and compose a new motion with the mini-clip(s) placed at the temporal spot(s) of interest. 
To perform this options, we use a binary mask $M\in\mathbb{B}^{T\times(JQ+C+3)}$ applied to the whole motion feature and not on specific joint-related features. 
As depicted in Fig.~\ref{fig:composition}, we use $M$ to select the region of interest (ROI) of the ``salsa dance" sequence, \ie $M$ values are ones for the frames that correspond to a clock-wise spin. 
Then, for option (a) we remove the ROI and inpaint the missing part of the motion as:
\begin{equation}
    T^{inpaint} = (M\odot \hat{T^{1}_{2}}) \oplus (\tilde{M}\odot \downarrow T^{salsa}),
\end{equation} 
where $\tilde{M}$ is the inverse of binary mask $M$ and $\hat{T^{1}_{2}} = G^{1}_{2}(G^{1}_{1}(z_{1}), z_{2})$ is the generated motion that is ``inpainted" in the downsampled ($\downarrow$) salsa sequence. The result of the ``inpainting" is presented in the middle of Fig.~\ref{fig:composition}. For option (b) we use the ROI as a standalone mini-clip $T^{ROI}$ which we downsample to the $L_{2}$ input level and concatenate a predefined numbers of ROIs with generated motion features from $L_{1}$ level. For example, as depicted in Fig.~\ref{fig:composition}(bottom), two $T^{ROI}$ - each representing a spin - are concatenated with $\hat{T}^{1}_{2}$ generated by generators trained with the ``salsa dance" sample. 
The two spins are smoothly blended into the generated salsa dance sequence in the desired time steps.

\begin{figure}
    \centering
    \includegraphics[width=\columnwidth]{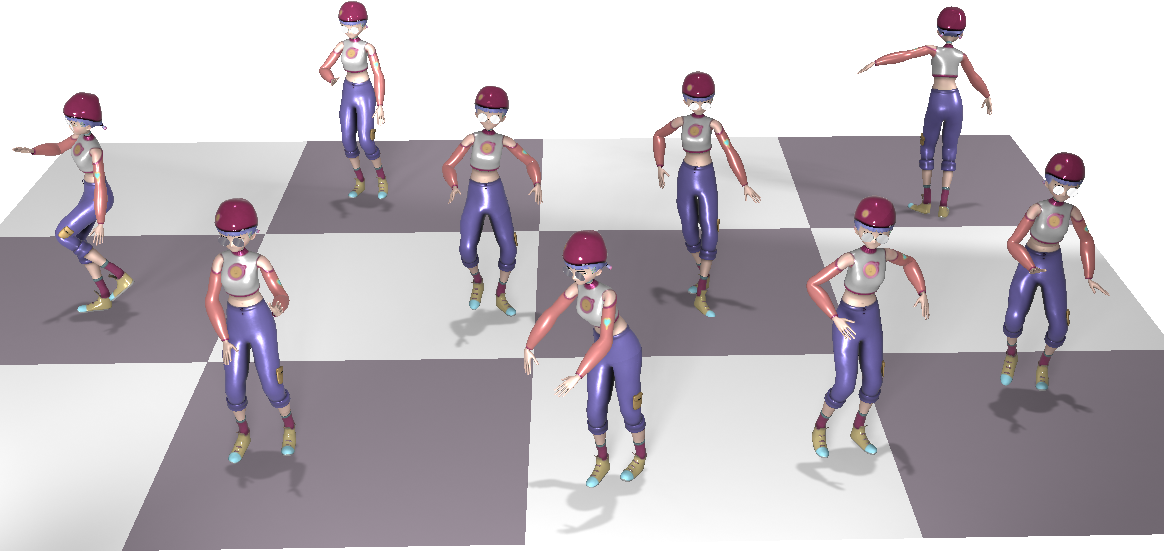}
    \caption{An exampled of crowd generation using the model training on the ``dancing" Mixamo benchmark sample.}
    \label{fig:crowd}
\end{figure}

\textbf{Crowd generation \& motion expansion.}~Single-shot learning enables the generation of motions with common low-frequency features, \ie same motion base, and small variations in the high-level features. 
This means that by sampling multiple codes from a Gaussian distribution $\mathcal{N}(0, I)$, we can generate a crowd performing similar motions. 
An example of crowd generation is depicted in Fig.~\ref{fig:crowd}. Another straight-forward application is the motion expansion. 
Since the skeleton-aware convolutions can be applied to a motion feature of arbitrary size, we can concatenate generated features $\hat{T}^{2}_{4}$ to the downsampled original motion features $T^{2}_{4}$ at the temporal dimension and use them as input to the corresponding $G_{\{2,\dots,S\}}$. 

\begin{figure}
    \centering
    \includegraphics[width=\columnwidth]{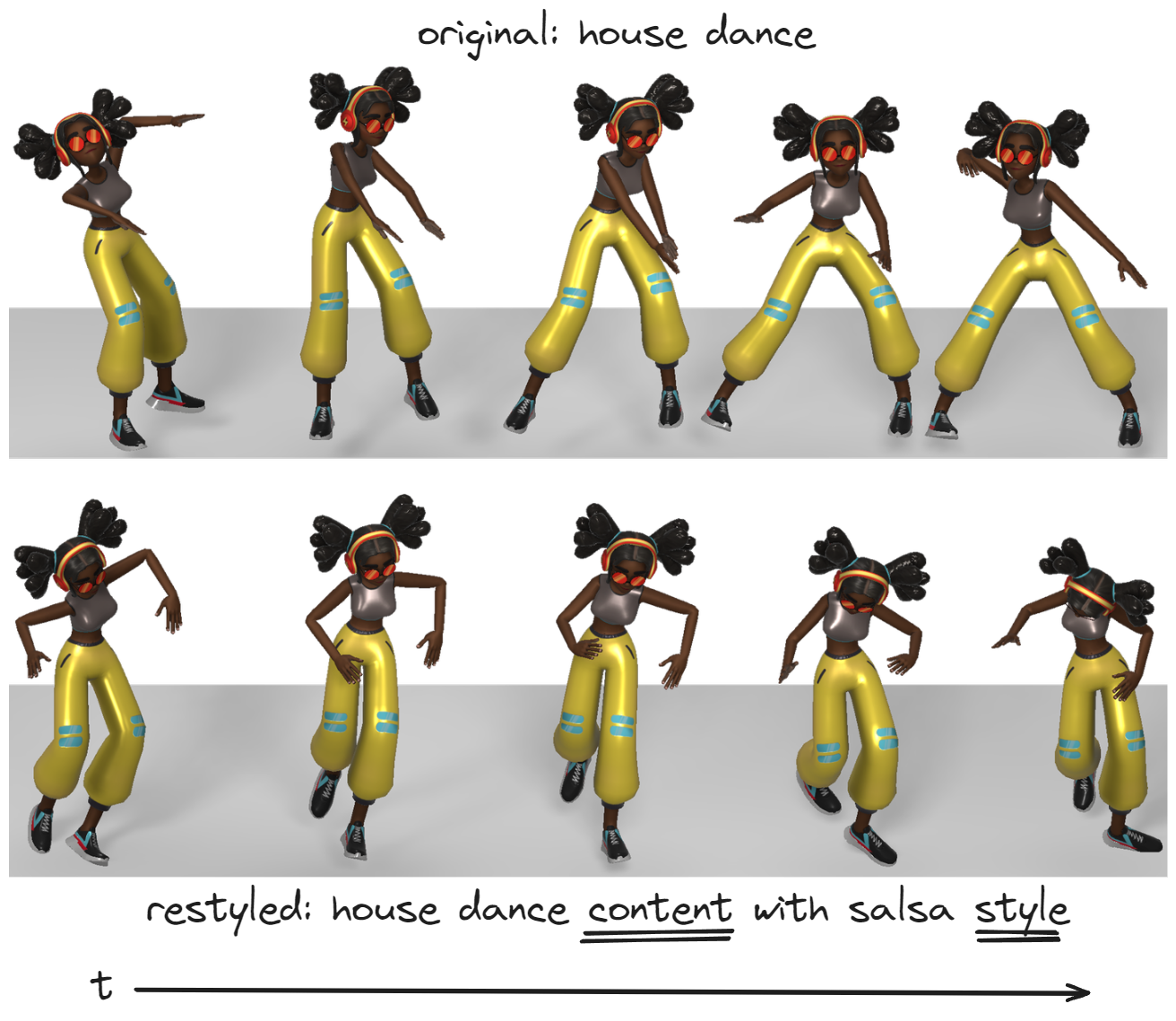}
    \caption{Restyling ``house dancing" (top) using a generator trained on a ``salsa dancing" Mixamo benchmark sample (bottom). Hands pose and knees proximity are characteristics of the ``salsa dancing" style.}
    \label{fig:restyle}
\end{figure}

\textbf{Re-styling.}~This has been the most discussed application in single-shot generation as it is relatively trivial for the image domain, however applying style on a motion is challenging. 
In the image domain, transferring style is the process of applying texture encoded information on image content (\eg applying a style of another artist on a painting as in~\cite{gatys2016style}). 
In the motion domain, style transfer is realized as applying high-frequency details on low-frequency features that correspond to a certain motion. 
This means that one cannot apply a dancing style from a stationary (\ie minor translation) motion to a walking one with single-shot generation. 
However, even with some restrictions, re-styling a motion in real-time is still valuable and leads to interesting results, as the example in Fig.~\ref{fig:restyle}. 
To re-style motion $T_{x}$ with style from motion $T_{y}$, we use the generators $G^{y}_{i}$ with $i \in {2, \dots S}$, \ie the stages that learn the high-level features (style) of $T_{y}$ with a downsampled version of $T_{x}$ that corresponds to the content ($T^{C}_{x}$). 
Note that the temporal downsampling process operates as low-pass filter, encoding the coarsest features of $T_{x}$ as its content.

\section{Conclusion}
\label{sec:conclusion}

In this work we investigate the performance of the single-sample generation GANs and address its key challenges. 
Building on prior work for learning motion generation from a single sample, we propose an loss weight annealing technique for enabling mini-batch training without compromising the adversarial equilibrium. 
To further minimize the required training iterations we propose a certain cross-stage weight initialization process based on a statistical analysis that exposes correlations between GAN stages. 
Overall, having similar performance in quality and diversity as anchor, our GAN improves GANimator and SinMDM train time and achieves impressive results in real-time applications without the need for re-training. 
Next steps include the integration of prior knowledge to further speed up training performance without compromising quality or diversity, as well as the investigation for a more specialized metric that will combine coverage with quality to be able to indicate when we sacrifice tail mini-motions for generated motion smoothness or variation.\\

\noindent\textbf{Acknowledgements.} This work was supported by EU's Horizon Europe Programme project EMIL-XR [GA 101070533].

\newpage
{
    \small
    \bibliographystyle{ieeenat_fullname}
    \bibliography{main}

\begin{thebibliography}{50}
\providecommand{\natexlab}[1]{#1}
\providecommand{\url}[1]{\texttt{#1}}
\expandafter\ifx\csname urlstyle\endcsname\relax
  \providecommand{\doi}[1]{doi: #1}\else
  \providecommand{\doi}{doi: \begingroup \urlstyle{rm}\Url}\fi

\bibitem[Aberman et~al.(2020)Aberman, Li, Lischinski, Sorkine-Hornung, Cohen-Or, and Chen]{aberman2020skeleton}
Kfir Aberman, Peizhuo Li, Dani Lischinski, Olga Sorkine-Hornung, Daniel Cohen-Or, and Baoquan Chen.
\newblock Skeleton-aware networks for deep motion retargeting.
\newblock \emph{ACM Trans. Graph. (TOG)}, 39\penalty0 (4):\penalty0 62, 2020.

\bibitem[Arar et~al.(2022)Arar, Shamir, and Bermano]{arar2022cvpr}
Moab Arar, Ariel Shamir, and Amit~H. Bermano.
\newblock Learned queries for efficient local attention.
\newblock In \emph{Proc. IEEE Conf. Comput. Vis. Pattern Recog. (CVPR)}, pages 10841--10852, 2022.

\bibitem[Arikan and Forsyth(2002)]{arikan2002interactive}
Okan Arikan and David~A Forsyth.
\newblock Interactive motion generation from examples.
\newblock \emph{ACM Trans. Graph. (TOG)}, 21\penalty0 (3):\penalty0 483--490, 2002.

\bibitem[Arikan et~al.(2003)Arikan, Forsyth, and O'Brien]{arikan2003motion}
Okan Arikan, David~A Forsyth, and James~F O'Brien.
\newblock Motion synthesis from annotations.
\newblock In \emph{Proc. ACM SIGGRAPH}, pages 402--408. 2003.

\bibitem[Athanasiou et~al.(2022)Athanasiou, Petrovich, Black, and Varol]{athanasiou2022teach}
Nikos Athanasiou, Mathis Petrovich, Michael~J Black, and G{\"u}l Varol.
\newblock T{EACH}: {T}emporal action composition for 3{D} humans.
\newblock In \emph{Proc. Int. Conf. 3D Vis. (3DV)}, pages 414--423, 2022.

\bibitem[Azadi et~al.(2023)Azadi, Shah, Hayes, Parikh, and Gupta]{azadi2023make}
Samaneh Azadi, Akbar Shah, Thomas Hayes, Devi Parikh, and Sonal Gupta.
\newblock Make-an-animation: {L}arge-scale text-conditional 3{D} human motion generation.
\newblock In \emph{Proc. IEEE Conf. Comput. Vis. Pattern Recog. (CVPR)}, pages 15039--15048, 2023.

\bibitem[Bergmann et~al.(2017)Bergmann, Jetchev, and Vollgraf]{bergmann2017icml}
Urs Bergmann, Nikolay Jetchev, and Roland Vollgraf.
\newblock Learning texture manifolds with the periodic spatial {GAN}.
\newblock In \emph{Proc. Int. Conf. Mach. Learn. (ICML)}, pages 469--477, 2017.

\bibitem[Bowden(2000)]{bowden2000learning}
Richard Bowden.
\newblock Learning statistical models of human motion.
\newblock In \emph{Proc. IEEE Conf. Comput. Vis. Pattern Recog. Worksh. (CVPRW)}, 2000.

\bibitem[Feng et~al.(2024)Feng, Lin, Dwivedi, Sun, Patel, and Black]{feng2023posegpt}
Yao Feng, Jing Lin, Sai~Kumar Dwivedi, Yu Sun, Priyanka Patel, and Michael~J. Black.
\newblock Chat{P}ose: {C}hatting about 3{D} human pose.
\newblock In \emph{Proc. IEEE Conf. Comput. Vis. Pattern Recog. (CVPR)}, 2024.

\bibitem[Gatys et~al.(2016)Gatys, Ecker, and Bethge]{gatys2016style}
Leon~A. Gatys, Alexander~S. Ecker, and Matthias Bethge.
\newblock Image style transfer using convolutional neural networks.
\newblock In \emph{Proc. IEEE Conf. Comput. Vis. Pattern Recog. (CVPR)}, pages 2414--2423, 2016.

\bibitem[Goodfellow et~al.(2020)Goodfellow, Pouget-Abadie, Mirza, Xu, Warde-Farley, Ozair, Courville, and Bengio]{goodfellow2020generative}
Ian Goodfellow, Jean Pouget-Abadie, Mehdi Mirza, Bing Xu, David Warde-Farley, Sherjil Ozair, Aaron Courville, and Yoshua Bengio.
\newblock Generative adversarial networks.
\newblock \emph{Communications of the ACM}, 63\penalty0 (11):\penalty0 139--144, 2020.

\bibitem[Granot et~al.(2022)Granot, Feinstein, Shocher, Bagon, and Irani]{granot2022drop}
Niv Granot, Ben Feinstein, Assaf Shocher, Shai Bagon, and Michal Irani.
\newblock Drop the {GAN}: {I}n defense of patches nearest neighbors as single image generative models.
\newblock In \emph{Proc. IEEE Conf. Comput. Vis. Pattern Recog. (CVPR)}, pages 13460--13469, 2022.

\bibitem[Gulrajani et~al.(2017)Gulrajani, Ahmed, Arjovsky, Dumoulin, and Courville]{gulrajani2017neurips}
Ishaan Gulrajani, Faruk Ahmed, Martin Arjovsky, Vincent Dumoulin, and Aaron~C Courville.
\newblock Improved training of {W}asserstein {GAN}s.
\newblock In \emph{Adv. Neural Inform. Process. Syst.}, 2017.

\bibitem[Gur et~al.(2020)Gur, Benaim, and Wolf]{gur2020hierarchical}
Shir Gur, Sagie Benaim, and Lior Wolf.
\newblock Hierarchical patch {VAE}-{GAN}: {G}enerating diverse videos from a single sample.
\newblock \emph{Adv. Neural Inform. Process. Syst.}, 33:\penalty0 16761--16772, 2020.

\bibitem[Hinz et~al.(2021)Hinz, Fisher, Wang, and Wermter]{hinz2021improved}
Tobias Hinz, Matthew Fisher, Oliver Wang, and Stefan Wermter.
\newblock Improved techniques for training single-image {GAN}s.
\newblock In \emph{Proc. IEEE Win. Conf. on App. Comput. Vis. (WACV)}, pages 1300--1309, 2021.

\bibitem[Ho et~al.(2020{\natexlab{a}})Ho, Jain, and Abbeel]{ho2020ddpm}
Jonathan Ho, Ajay Jain, and Pieter Abbeel.
\newblock Denoising diffusion probabilistic models.
\newblock In \emph{Adv. Neural Inform. Process. Syst.}, page 6840–6851, 2020{\natexlab{a}}.

\bibitem[Ho et~al.(2020{\natexlab{b}})Ho, Jain, and Abbeel]{ho2020denoising}
Jonathan Ho, Ajay Jain, and Pieter Abbeel.
\newblock Denoising diffusion probabilistic models.
\newblock \emph{Adv. Neural Inform. Process. Syst.}, 33:\penalty0 6840--6851, 2020{\natexlab{b}}.

\bibitem[Holden et~al.(2016)Holden, Saito, and Komura]{holden2016deep}
Daniel Holden, Jun Saito, and Taku Komura.
\newblock A deep learning framework for character motion synthesis and editing.
\newblock \emph{ACM Trans. Graph. (TOG)}, 35\penalty0 (4):\penalty0 1--11, 2016.

\bibitem[Isola et~al.(2017)Isola, Zhu, Zhou, and Efros]{isola2017cvpr}
Phillip Isola, Jun\-Yan Zhu, Tinghui Zhou, and Alexei~A. Efros.
\newblock Learned queries for efficient local attention.
\newblock In \emph{Proc. IEEE Conf. Comput. Vis. Pattern Recog. (CVPR)}, pages 1125--1134, 2017.

\bibitem[Jetchev et~al.(2016)Jetchev, Bergmann, and Vollgraf]{jetchev2016nipsw}
Nikolay Jetchev, Urs Bergmann, and Roland Vollgraf.
\newblock Texture synthesis with spatial generative adversarial networks. workshop on adversarial training.
\newblock In \emph{Adv. Neural Inform. Process. Syst. Worksh.}, 2016.

\bibitem[Jiang et~al.(2023)Jiang, Chen, Liu, Yu, Yu, and Chen]{jiang2023motiongpt}
Biao Jiang, Xin Chen, Wen Liu, Jingyi Yu, Gang Yu, and Tao Chen.
\newblock Motion{GPT}: {H}uman motion as a foreign language.
\newblock pages 20067--20079, 2023.

\bibitem[Kim et~al.(2023)Kim, Kim, and Choi]{kim2022flame}
Jihoon Kim, Jiseob Kim, and Sungjoon Choi.
\newblock F{LAME}: {F}ree-form language-based motion synthesis \& editing.
\newblock pages 8255--8263, 2023.

\bibitem[Kornblith et~al.(2019)Kornblith, Norouzi, Lee, and Hinton]{kornblith2019similarity}
Simon Kornblith, Mohammad Norouzi, Honglak Lee, and Geoffrey Hinton.
\newblock Similarity of neural network representations revisited.
\newblock In \emph{Proc. Int. Conf. Mach. Learn. (ICML)}, pages 3519--3529, 2019.

\bibitem[Kovar et~al.(2023)Kovar, Gleicher, and Pighin]{kovar2023motion}
Lucas Kovar, Michael Gleicher, and Fr{\'e}d{\'e}ric Pighin.
\newblock Motion graphs.
\newblock In \emph{Seminal Graphics Papers: Pushing the Boundaries, Volume 2}, pages 723--732. 2023.

\bibitem[Kulikov et~al.(2023)Kulikov, Yadin, Kleiner, and Michaeli]{kulikov2023sinddm}
Vladimir Kulikov, Shahar Yadin, Matan Kleiner, and Tomer Michaeli.
\newblock Sin{DDM}: {A} single image denoising diffusion model.
\newblock In \emph{Proc. Int. Conf. Mach. Learn. (ICML)}, pages 17920--17930, 2023.

\bibitem[Li and Wand(2016)]{li2016patch}
Chuan Li and Michael Wand.
\newblock Precomputed real-time texture synthesis with {M}arkovian generative adversarial networks.
\newblock In \emph{Proc. Eur. Conf. Comput. Vis. (ICCV)}, pages 702--716, 2016.

\bibitem[Li et~al.(2023)Li, Li, Savarese, and Hoi]{blip2}
Junnan Li, Dongxu Li, Silvio Savarese, and Steven Hoi.
\newblock B{LIP-2}: {B}ootstrapping language-image pre-training with frozen image encoders and large language models.
\newblock In \emph{Proc. Int. Conf. Mach. Learn. (ICML)}, pages 19730--19742, 2023.

\bibitem[Li et~al.(2022)Li, Aberman, Zhang, Hanocka, and Sorkine-Hornung]{li2022ganimator}
Peizhuo Li, Kfir Aberman, Zihan Zhang, Rana Hanocka, and Olga Sorkine-Hornung.
\newblock G{AN}imator: {N}eural motion synthesis from a single sequence.
\newblock \emph{ACM Trans. Graph. (TOG)}, 41\penalty0 (4):\penalty0 138, 2022.

\bibitem[Li et~al.(2002)Li, Wang, and Shum]{li2002motion}
Yan Li, Tianshu Wang, and Heung-Yeung Shum.
\newblock Motion texture: {A} two-level statistical model for character motion synthesis.
\newblock In \emph{Proc. Annual Conf. on Comp. Graph. and Inter. Tech. (CGIT)}, pages 465--472, 2002.

\bibitem[Nichol et~al.()Nichol, Dhariwal, Ramesh, Shyam, Mishkin, McGrew, Sutskever, and Chen]{nichol2022icml}
Alexander~Quinn Nichol, Prafulla Dhariwal, Aditya Ramesh, Pranav Shyam, Pamela Mishkin, Bob McGrew, Ilya Sutskever, and Mark Chen.
\newblock {GLIDE:} {T}owards photorealistic image generation and editing with text-guided diffusion models.
\newblock In \emph{Proc. Int. Conf. Mach. Learn. (ICML)}, pages 16784--16804.

\bibitem[Paszke et~al.(2019)Paszke, Gross, Massa, Lerer, Bradbury, Chanan, Killeen, Lin, Gimelshein, Antiga, et~al.]{paszke2019pytorch}
Adam Paszke, Sam Gross, Francisco Massa, Adam Lerer, James Bradbury, Gregory Chanan, Trevor Killeen, Zeming Lin, Natalia Gimelshein, Luca Antiga, et~al.
\newblock Py{T}orch: {A}n imperative style, high-performance deep learning library.
\newblock \emph{Adv. Neural Inform. Process. Syst.}, 32, 2019.

\bibitem[Petrovich et~al.(2022)Petrovich, Black, and Varol]{petrovich2022temos}
Mathis Petrovich, Michael~J Black, and G{\"u}l Varol.
\newblock T{EMOS}: {G}enerating diverse human motions from textual descriptions.
\newblock In \emph{Proc. Eur. Conf. Comput. Vis. (ICCV)}, pages 480--497, 2022.

\bibitem[Raab et~al.(2024)Raab, Leibovitch, Tevet, Arar, Bermano, and Cohen-Or]{raab2024single}
Sigal Raab, Inbal Leibovitch, Guy Tevet, Moab Arar, Amit~H Bermano, and Daniel Cohen-Or.
\newblock Single motion diffusion.
\newblock In \emph{Proc. Int. Conf. Learn. Represent. (ICLR)}, 2024.

\bibitem[Rott~Shaham et~al.(2019)Rott~Shaham, Dekel, and Michaeli]{rottshaham2019singan}
Tamar Rott~Shaham, Tali Dekel, and Tomer Michaeli.
\newblock Sin{GAN}: {L}earning a generative model from a single natural image.
\newblock In \emph{Proc. Int. Conf. Comput. Vis. (ICCV)}, 2019.

\bibitem[Salimans et~al.(2016)Salimans, Goodfellow, Zaremba, Cheung, Radford, Chen, and Chen]{salimans2016neurips}
Tim Salimans, Ian Goodfellow, Wojciech Zaremba, Vicki Cheung, Alec Radford, Xi Chen, and Xi Chen.
\newblock Improved techniques for training {GAN}s.
\newblock In \emph{Adv. Neural Inform. Process. Syst.}, 2016.

\bibitem[Shafir et~al.(2023)Shafir, Tevet, Kapon, and Bermano]{shafir2023human}
Yoni Shafir, Guy Tevet, Roy Kapon, and Amit~Haim Bermano.
\newblock Human motion diffusion as a generative prior.
\newblock In \emph{Proc. Int. Conf. Learn. Represent. (ICLR)}, 2023.

\bibitem[Shocher et~al.(2018)Shocher, Cohen, and Irani]{shocher2018zero}
Assaf Shocher, Nadav Cohen, and Michal Irani.
\newblock ``{Z}ero-shot" super-resolution using deep internal learning.
\newblock In \emph{Proc. IEEE Conf. Comput. Vis. Pattern Recog. (CVPR)}, pages 3118--3126, 2018.

\bibitem[Shocher et~al.(2019)Shocher, Bagon, Isola, and Irani]{shocher2019ingan}
Assaf Shocher, Shai Bagon, Phillip Isola, and Michal Irani.
\newblock In{GAN}: {C}apturing and retargeting the ``{DNA}"" of a natural image.
\newblock In \emph{Proc. Int. Conf. Comput. Vis. (ICCV)}, pages 4491--4500, 2019.

\bibitem[Son et~al.(2023)Son, Park, Guibas, and Wetzstein]{son2023singraf}
Minjung Son, Jeong~Joon Park, Leonidas Guibas, and Gordon Wetzstein.
\newblock Sin{GRAF}: {L}earning a 3{D} generative radiance field for a single scene.
\newblock In \emph{Proc. IEEE Conf. Comput. Vis. Pattern Recog. (CVPR)}, pages 8507--8517, 2023.

\bibitem[Sushko et~al.(2021)Sushko, Gall, and Khoreva]{sushko2021one}
Vadim Sushko, Jurgen Gall, and Anna Khoreva.
\newblock One-shot {GAN}: {L}earning to generate samples from single images and videos.
\newblock In \emph{Proc. IEEE Conf. Comput. Vis. Pattern Recog. (CVPR)}, pages 2596--2600, 2021.

\bibitem[Szegedy et~al.(2015)Szegedy, Liu, Jia, Sermanet, Reed, Anguelov, Erhan, Vanhoucke, and Rabinovich]{szegedy2015inception}
Christian Szegedy, Wei Liu, Yangqing Jia, Pierre Sermanet, Scott Reed, Dragomir Anguelov, Dumitru Erhan, Vincent Vanhoucke, and Andrew Rabinovich.
\newblock Going deeper with convolutions.
\newblock In \emph{Proc. IEEE Conf. Comput. Vis. Pattern Recog. (CVPR)}, pages 1--9, 2015.

\bibitem[Tevet et~al.()Tevet, Raab, Gordon, Shafir, Cohen-or, and Bermano]{tevet2023human}
Guy Tevet, Sigal Raab, Brian Gordon, Yoni Shafir, Daniel Cohen-or, and Amit~Haim Bermano.
\newblock Human motion diffusion model.
\newblock In \emph{Proc. Int. Conf. Learn. Represent. (ICLR)}.

\bibitem[Wu et~al.(2023)Wu, Liu, Vondrick, and Zheng]{wu2023sin3dm}
Rundi Wu, Ruoshi Liu, Carl Vondrick, and Changxi Zheng.
\newblock Sin3{DM}: {L}earning a diffusion model from a single 3{D} textured shape.
\newblock In \emph{Proc. Int. Conf. Learn. Represent. (ICLR)}, 2023.

\bibitem[Yoo and Chen(2021)]{yoo2021sinir}
Jihyeong Yoo and Qifeng Chen.
\newblock Sin{IR}: {E}fficient general image manipulation with single image reconstruction.
\newblock In \emph{Proc. Int. Conf. Mach. Learn. (ICML)}, pages 12040--12050, 2021.

\bibitem[Zhang et~al.(2023)Zhang, Li, and Bing]{damonlpsg2023videollama}
Hang Zhang, Xin Li, and Lidong Bing.
\newblock Video-{LL}a{M}a: {A}n instruction-tuned audio-visual language model for video understanding.
\newblock 2023.

\bibitem[Zhang et~al.(2018)Zhang, Isola, Efros, Shechtman, and Wang]{zhang2018cvpr}
R. Zhang, P. Isola, A.~A. Efros, E. Shechtman, and O. Wang.
\newblock The unreasonable effectiveness of deep features as a perceptual metric.
\newblock In \emph{Proc. IEEE Conf. Comput. Vis. Pattern Recog. (CVPR)}, pages 586--595, 2018.

\bibitem[Zhang et~al.(2021)Zhang, Han, and Guo]{zhang2021exsingan}
ZiCheng Zhang, CongYing Han, and TianDe Guo.
\newblock Ex{S}in{GAN}: {L}earning an explainable generative model from a single image.
\newblock \emph{arXiv preprint arXiv:2105.07350}, 2021.

\bibitem[Zhang et~al.(2022)Zhang, Liu, Han, Shi, Guo, and Zhou]{zhang2022petsgan}
Zicheng Zhang, Yinglu Liu, Congying Han, Hailin Shi, Tiande Guo, and Bowen Zhou.
\newblock Pets{GAN}: {R}ethinking priors for single image generation.
\newblock In \emph{Proc. AAAI}, pages 3408--3416, 2022.

\bibitem[Zhao et~al.(2024)Zhao, Chen, Yang, Liu, Deng, Cai, Wang, Yin, and Du]{llm2024explain}
Haiyan Zhao, Hanjie Chen, Fan Yang, Ninghao Liu, Huiqi Deng, Hengyi Cai, Shuaiqiang Wang, Dawei Yin, and Mengnan Du.
\newblock Explainability for large language models: {A} survey.
\newblock \emph{ACM Trans. Intell. Syst. Technol.}, 15\penalty0 (2), 2024.

\bibitem[Zhou et~al.(2018)Zhou, Zhu, Bai, Lischinski, Cohen-Or, and Huang]{zhou2018non}
Yang Zhou, Zhen Zhu, Xiang Bai, Dani Lischinski, Daniel Cohen-Or, and Hui Huang.
\newblock Non-stationary texture synthesis by adversarial expansion.
\newblock \emph{arXiv preprint arXiv:1805.04487}, 2018.

\end{thebibliography}
}

\end{document}